\documentclass{article}
\usepackage{ijcai24}

\usepackage{times}
\usepackage{soul}
\usepackage{tabularx}
\usepackage{color}
\usepackage{url}
\usepackage{multirow}
\usepackage{hyperref}
\hypersetup {hidelinks}
\usepackage[utf8]{inputenc}
\usepackage[small]{caption}
\usepackage{graphicx}
\usepackage{amsmath}
\usepackage{amsthm}
\usepackage{amssymb}
\usepackage{booktabs}
\usepackage{caption}
\usepackage{xcolor}
\usepackage[linesnumbered,ruled,vlined]{algorithm2e}
\usepackage[switch]{lineno}
\SetKwInput{KwInput}{Input}                % Set the Input
\SetKwInput{KwOutput}{Output}              % set the Output

\title{TFWT: Tabular Feature Weighting with Transformer}

\author{
Xinhao Zhang$^1$
\and
Zaitian Wang$^{2,3}$,
Lu Jiang$^4$,
Wanfu Gao$^5$,
Pengfei Wang$^{2,3}$\footnotemark[1]\And
Kunpeng Liu$^1$\footnotemark[1]
\affiliations
$^1$Portland State University\\
$^2$Computer Network Information Center, Chinese Academy of Sciences\\
$^3$University of Chinese Academy of Sciences, Chinese Academy of Sciences\\
$^4$Dalian Maritime University\\
$^5$Jilin University
\emails
xinhaoz@pdx.edu,
wangzaitian23@mails.ucas.ac.cn,
jiangl761@dlmu.edu.cn,
gaowf@jlu.edu.cn,
pfwang@cnic.cn,
kunpeng@pdx.edu
}
\begin{document}

\maketitle
\setcounter{footnote}{0}
\renewcommand{\thefootnote}{\fnsymbol{footnote}}
% \footnotetext[1]{Both authors contributed equally to this research}
\footnotetext[1]{Corresponding authors.}
\setcounter{footnote}{0}
\renewcommand{\thefootnote}{\arabic{footnote}}

\begin{abstract}
In this paper, we propose a novel feature weighting method to address the limitation of existing feature processing methods for tabular data. Typically the existing methods assume equal importance across all samples and features in one dataset. This simplified processing methods overlook the unique contributions of each feature, and thus may miss important feature information. As a result, it leads to suboptimal performance in complex datasets with rich features. To address this problem, we introduce \textbf{\underline{T}}abular \textbf{\underline{F}}eature \textbf{\underline{W}}eighting with \textbf{\underline{T}}ransformer, a novel feature weighting approach for tabular data. Our method adopts Transformer to capture complex feature dependencies and contextually assign appropriate weights to discrete and continuous features. Besides, we employ a reinforcement learning strategy to further fine-tune the weighting process. Our extensive experimental results across various real-world datasets and diverse downstream tasks show the effectiveness of \textbf{TFWT} and highlight the potential for enhancing feature weighting in tabular data analysis.
\end{abstract}

\section{Introduction}
Extracting feature information from data is one of the most crucial tasks in machine learning, especially for classification and prediction tasks~\cite{bishop1995neural}. Different features play varied roles in data representation and pattern recognition, thus directly impacting the model's learning efficiency and prediction accuracy. Effective feature engineering can enhance a model's ability to handle complex data significantly and help capture critical information in the data.

\begin{figure}[t]
  \centering
  \includegraphics[width=1\linewidth]{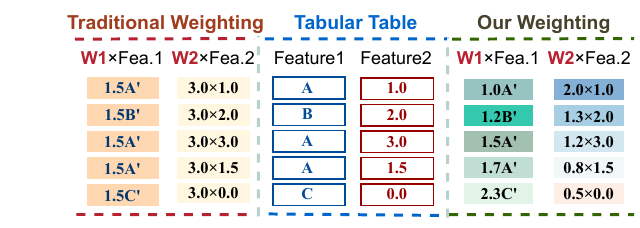}
  % \vspace{-6mm}
  \caption{A Demonstration of feature weighting. Traditional feature weighting methods assign the same weight to one feature. Our weighting method assigns different weights to different samples in one feature.}
  % \vspace{-4mm}
  \label{fig:feature_weighting}
\end{figure}

In feature engineering, a traditional and common assumption is that each feature in a dataset is equally essential. Treating all features equally important does simplify the process, but it ignores the fact that each feature contributes uniquely to the downstream task~\cite{daszykowski2007robust}. Some features with rich information may significantly impact the model outcome, while others may contribute less or even introduce noise and mislead the model~\cite{garcia2015data}. Treating all features with equal weight may dilute the information of important features and cause less important features to over-influence the model, thus diminishing the overall effectiveness of information extraction.
% This limitation becomes more apparent when dealing with complex data with rich features.
Feature weighting is a technique that assigns weights to each feature in the dataset. The main goal of feature weighting is to optimize the feature space by assigning weights to each feature according to feature importance, thus enhancing the model performance. Feature weighting methods can be categorized based on learning strategies and weighting methods. Supervised feature weighting uses actual data labels to determine the weights of features~\cite{chen2017feature,nino2020analysis,wang2022semi}. 
% In global supervised feature weighting, each feature is assigned a single weight based on its overall relevance across the entire dataset, thus capturing the general importance of features. Local supervised feature weighting assigns multiple weights to a single feature, which reflects its varying relevance across different subsets of data or classes. 
Unsupervised feature weighting utilizes the intrinsic characteristics of the dataset to assign weight without relying on label information~\cite{zhang2018tw}.

Feature weighting can also be classified based on feature types or feature information. Weighting by feature types focuses on the inherent properties of features, like discreteness or continuity. This approach is often influenced by data structure~\cite{zhou2021feature,xue2023external,hashemzadeh2019new,cardie1997improving,lee2011calculating}. Conversely, weighting by feature information emphasizes the informational content of features. It assesses the importance of features based on their correlation and contribution to the predictive model, typically using statistical measures or machine learning algorithms~\cite{kang2019few,liu2004text,druck2008learning,WinGNN,liu2019automating}. However, these methods do not effectively capture the complex relationships between features and cause risks like overfitting, local optimality, and noise sensitivity. Moreover, as illustrated in Figure~\ref{fig:feature_weighting}, these methods assign the same weight to every sample (row) of each feature, rather than assigning personalized weights to different samples.

% To tackle these challenges, we propose a novel approach based on Transformer. Transformer is well-known for its exceptional performance in natural language processing. It acts as a ``translator'' that converts one language into another with self-attention mechanisms~\cite{vaswani2017attention,radford2019language,luong2015effective,ZHOU2023109203,9206016}. In this study, we leverage this characteristic to ``translate'' appropriate weights for features in tabular data.

In addressing the limitations of traditional feature weighting methods, we adopt a new approach based on the Transformer model. One core feature of the Transformer is its self-attention mechanism~\cite{vaswani2017attention,radford2019language,luong2015effective,ZHOU2023109203,9206016,hashemzadeh2019new}. This mechanism can effectively identify complex dependencies and interactions among data features. In feature weighting, the self-attention mechanism assigns attention weights by considering the relevance and contribution of features, enabling the model to learn and adapt to specific dataset patterns during the training process. Thus, our Transformer-based approach can assign higher weights to features that significantly influence the model output. In this way, the model can focus on the most critical information. With self-attention, the Transformer can effectively capture the contextual information and inter-feature relationships within the tabular data.

The Transformer also employs a multi-head attention mechanism operated with several self-attention components operating in parallel. Each attention head focuses on different aspects of the dataset, and captures diverse patterns and dependencies. Thus the model can understand the feature relationships from multiple perspectives. This application of multi-head attention significantly enhances the model's capability in determining feature weights. Thus, by integrating self-attention and multi-head attention mechanisms, our Transformer-based method effectively identifies and assigns feature weights. It adapts to various data patterns and complicated relationships among features.
% This approach significantly enhances the efficiency and accuracy of feature weighting.

\par To further stabilize and enhance this weighting structure, we need an effective fine-tuning method. Adopting reinforcement learning~\cite{schulman2017proximal,fan2020autofs} for fine-tuning is a common strategy~\cite{ziegler2019fine,fickinger2021scalable,ouyang2022training,DBLP:journals/ijon/ZhuLQSCN20}. Notably, the Proximal Policy Optimization (PPO) network~\cite{schulman2017proximal} has the advantage of stability and efficiency in fine-tuning by enhancing policies while ensuring stable updates~\cite{zhu2021recommending}. Specifically, the PPO network fits the task of reducing information redundancy within the data. In this scenario, information redundancy refers to the presence of repetitive or irrelevant information. Redundant features lead to possible overfitting in training. By reducing redundancy, the data becomes more concise and focused on the most informative features. By minimizing redundancy, learning becomes more stable and focused, which helps to decrease classification variance. 

\par In summary, we propose a novel feature weighting method aiming to tackle several challenges: first, how to generate appropriate weights of features; second, how to evaluate the effectiveness of feature weights; and finally, how to fine-tune the feature weights according to the feedback from downstream tasks. Hence, we introduce a Transformer-enhanced feature weighting framework in response to the challenges outlined. This framework leverages the strength of the Transformer architecture to assign weights to features by capturing intricate contextual relationships among these features. We evaluate the effectiveness of feature weighting by the improvement of downstream task's performance.
% By integrating these contextually weighted features into downstream models, we observe a notable boost in several performance metrics. 
Further, we adopt a reinforcement learning strategy to fine-tune the output and reduce information variance. This adjustment enhances the model's stability and reliability.

Our contributions are summarized as follows:

\begin{itemize}
    \item We propose a novel feature weighting method for tabular data based on Transformer called \textbf{TFWT}. This new method can capture the dependencies between features with Transformer's attention mechanism to assign and adjust weights for features according to the feedback of downstream tasks. 

    \item We propose a fine-tuning method for the weighting process to further enhance the performance. This fine-tuning method adopts a reinforcement learning strategy, reducing the data information redundancy and classification variance. 
    \item We conduct extensive experiments and show that \textbf{TFWT} achieves significant performance improvements under varying datasets and downstream tasks, comparing with raw classifiers and baseline models. The experiments also show the effectiveness of fine-tuning process in reducing redundancy.

\end{itemize}

\section{Related Work}

\subsection{Feature Weighting}
Feature weighting, vital for enhancing machine learning, includes several approaches~\cite{CHEN20153142,CHEN2017340,CHOWDHURY2023109314,wang2004improving,yeung2002improving}.~\cite{liu2004text},~\cite{druck2008learning}, and~\cite{raghavan2006active} explored feedback integration, model constraints, and active learning enhancement. ~\cite{wang2013active} proposed an active SVM method for image retrieval. Techniques like weighted bootstrapping~\cite{barbe1995weighted}, chi-squared tests, TabTransformer~\cite{huang2020tabtransformer}, and cost-sensitive learning adjust weights through feature changes. These methods have limitations like overfitting or ignoring interactions. Our study focuses on adaptable weight distribution and improvement through feedback.
% \subsection{Tabular Prediction}
% In the field of tabular prediction, there are many well-known algorithms~\cite{chen2016xgboost,ke2017lightgbm,dorogush2018catboost,chen2022danets,gorishniy2021revisiting,cholakov2022gatedtabtransformer}. 
\subsection{Transformer}
The Transformer architecture, introduced by~\cite{vaswani2017attention}, has revolutionized many fields including natural language processing. Instead of relying on recurrence like its predecessors, it utilizes self-attention mechanisms to capture dependencies regardless of their distance in the input data. This innovation has led to several breakthroughs in various tasks. For instance, BERT model~\cite{devlin2018bert,clark2019does}, built upon the Transformer, set new records in multiple NLP benchmarks. Later,~\cite{radford2019language} extended these ideas with GPT-2 and GPT-3~\cite{brown2020language}, demonstrating impressive language generation capabilities. Concurrently,~\cite{raffel2020exploring} proposed a unified text-to-text framework for NLP transfer learning, achieving state-of-the-art results across multiple tasks.

\section{Methodology}

\begin{figure*}[htb]
  \centering
  \includegraphics[width=\textwidth]{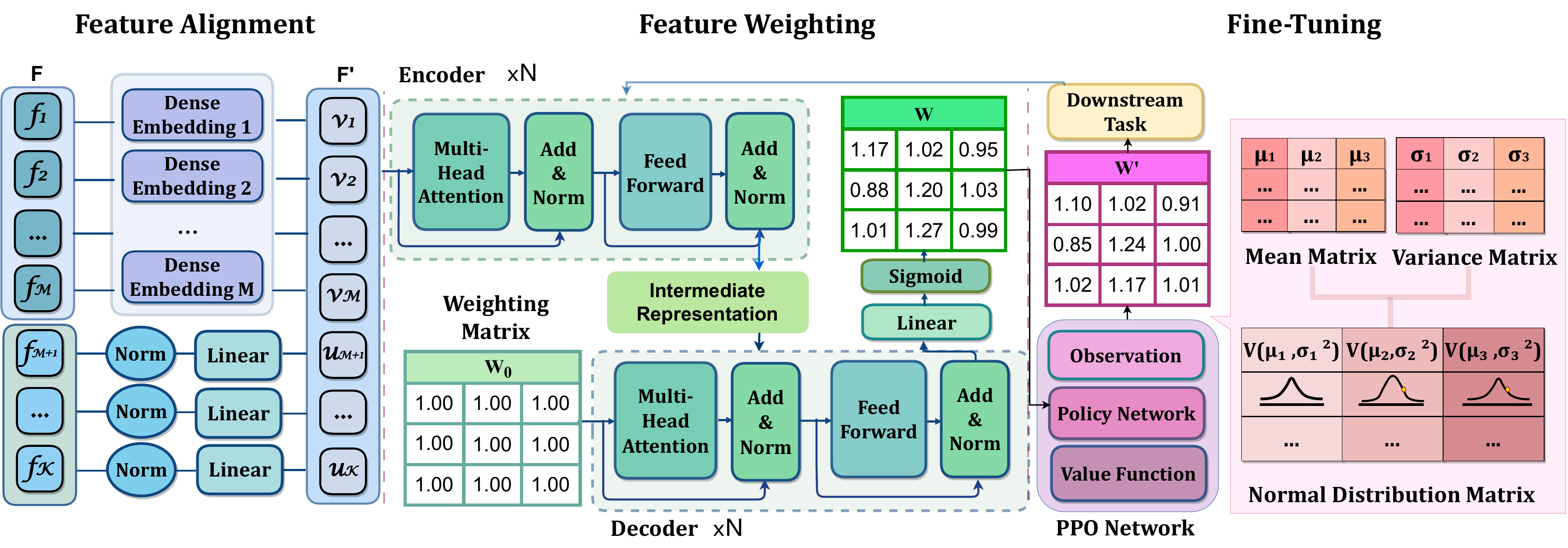}
  % \vspace{-5mm}
  \caption{
The framework consists of three components. In the alignment we convert discrete ($f_1$ to $f_M$) and continuous ($f_{M+1}$ to $f_K$) features into uniform-length vectors. In the weighting we initialize and reassign weights according to feature relationships. The fine-tuning process employs reinforcement learning to refine the weighting model.
  %Framework of Feature Weighting. (Left) The Fearture Alignment part and (Middle) the Feature Weighting part make the TFWT method an end-to-end architecture with (Right) Reinforcement Learning part to enhance it.
  }
  \label{fig:framework-feature-weighting}
  % \vspace{-3mm}
\end{figure*}
\subsection{Problem Formulation}
% We consider the problem setting of classification in $Y$ classes. Let $D = \{(\textbf{x}^{(i)}, y_i)\}_{i=1}^N$ be a $k$-dimensional dataset of $N$ samples, where $\textbf{x}^{(i)} \in \mathbb{R}^k$ is a sample and $y_i \in Y$ (discrete) is its label. The $j$-th feature of the $i$-th sample is represented as $f^{(i)}_j$, where the first $m$ features $f^{(i)}_j$ are discrete, and the remaining $n = k - m$ features are continuous. We define the dataset matrix as $X := [\textbf{x}^{(1)}, \ldots, \textbf{x}^{(N)}]^\top \in \mathbb{R}^{N \times k}$, and the labels as $Y := [y_1, \ldots, y_N]$.

We consider the problem setting of classification. Let $\mathcal{D} = \{\mathbf{F}, \mathbf{y}\}$ be a dataset with $K$ features and $N$ samples. We define the feature matrix $\mathbf{F} = \{\mathbf{f}_k\}_{k=1}^K$. We use $\mathbf{f}_k = \{f_k^1,\dots, f_k^i, \dots, f_k^N\}^\top$ to denote the $k$-th feature, where $f_k^i$ is the value of $i$-th sample on the $k$-th feature. $\mathbf{y}=[y_1, \ldots, y_N]^\top$ is the label vector. Without loss of generality, we assume the first $M$ features to be discrete, and the remaining $K - M$ features to be continuous. 
% We define the feature matrix as $\mathbf{F} = \{\mathbf{f}_k\}_{k=1}^K$.

In defining a weighting matrix $\mathbf{W} \in \mathbb{R}^{N \times K}$, each of whose elements corresponds to the elements of the feature matrix $\mathbf{F}$. This weighting matrix $\mathbf{W}$ is applied element-wisely to $\mathbf{F}$ to produce a weighted matrix $\mathbf{F}_{rew} = \mathbf{W} \odot \mathbf{F}$, where $\odot$ denotes the Hadamard product. In the weighting problem, we aim to find an optimized $\mathbf{W}$, so that $\mathbf{F}_{rew}$ can improve the downstream tasks' performance when substituting the original feature matrix $\mathbf{F}$ in predicting $\mathbf{y}$.

\subsection{Framework}
% In this paper,
We propose \textbf{TFWT}, a  \textbf{T}abular \textbf{F}eature \textbf{W}eighting with \textbf{T}ransformer method for tabular data. We aim to improve downstream tasks' performance by effectively incorporating the attention mechanism to capture the relations and interactions between features. To achieve this goal, we design a Transformer-based feature weighting pipeline with a fine-tuning strategy. As Figure~\ref{fig:framework-feature-weighting} shows, our method consists of three components: In the \textit{Feature Alignment}, we align different types of original features so that they are in the same space. In the \textit{Feature Weighting}, we encode the feature matrix to get its embedding via Transformer encoders, and then decode the embedding into feature weights. In the \textit{Fine-Tuning}, we design a reinforcement learning strategy to fine-tune the feature weights based on feedback from downstream tasks. 

% \begin{itemize}
%     \item \par In the \textbf{Feature Alignment}, we align different types of original features so that they are in the same space.
%     % numerically encode the dataset to ensure it is machine-readable.
%     \item \par In the \textbf{Feature Weighting}, we first encode the feature matrix to get its embedding via transformer encoder, and then decode the embedding into feature weights.
    
%     % we put the aligned feature matrix into a high-dimensional feature space, capturing profound relationships within the features. Then we utilize the attention mechanism, by training a Transformer model to learn from the encoded data and generate a weighting matrix.
%     \item \par In the \textbf{Fine-Tuning}, we design a reinforcement learning strategy to fine-tune the feature weights based on feedback from downstream tasks. 
% \end{itemize}
% \par Next, we will introduce each component and how we jointly optimize them in an end-to-end framework.

\subsection{Feature Alignment}

To effectively extract tabular data's features while maintaining a streamlined computation, we convert both discrete and continuous features into numerical vectors. 

\noindent \textbf{Discrete Feature Alignment.} We first encode the discrete features into numerical values. The encoded numerical values are then passed to a dense embedding layer,  transforming them into vectors for subsequent processes. For each discrete feature $\mathbf{f}_k$ ($k = 1, \ldots, M$), the encoded vector is:
\begin{equation}
\mathbf{v}_k = \text{Dense} (\mathbf{f}_k) .
\label{eq_discrete}
\end{equation}

\noindent \textbf{Continuous Feature Alignment.} We normalize all the continuous features with mean of 0 and variance of 1. We then design a linear layer to align their length with discrete features. For each continuous feature $\mathbf{f}_k$ ($k = M+1, \ldots, K$), the encoded vector is:
\begin{equation}
\mathbf{u}_k = \text{Linear}\left(\frac{\mathbf{f}_k - \mu_k}{\sigma_k}\right) ,
\label{eq_contious}
\end{equation}
where $\mu_k$ and $\sigma_k$ are the mean and standard deviation of the $k$-th feature, respectively.
Then the aligned feature matrix $\mathbf{F^{\prime}}$ is formed by concatenating these vectors:
\begin{equation}
\mathbf{F^{\prime}} = [\mathbf{v}_1, \ldots, \mathbf{v}_M, \mathbf{u}_{M+1}, \ldots, \mathbf{u}_K] .
\end{equation}

% The matrix is introduced to the Transformer's encoder in the feature weighting process, where the matrix is transformed into a latent representation with rich information. This representation will be used in both the feature weighting and the fine-tuning processes.
% \par During alignment, we embed the data matrix, now referred to as $D^{\prime}$. This step produces an embedded matrix $D_e$. During this embedding process, we transform each original feature $x$ in $D^{\prime}$ into an embedded vector using an embedding layer, which we denote as $e$. This transformation aims to capture and represent the detailed characteristics of each feature more informatively and efficiently.

\subsection{Feature Weighting}

Given aligned feature matrix $\mathbf{F^{\prime}}$, we aim to explore the relationships between features and assign proper feature weights. 
% to them.

% In this subsection, we first show how we identify and quantify the features within the data. Next, we discuss how we give a weighting matrix with decoding technology. After that, we show how we adjust the model with downstream tasks' feedback and back-propagate.
\noindent \textbf{Data Encoding.} To enhance the model's understanding and extract latent patterns and relations from the data, we put $\mathbf{F^{\prime}}$ into the encoders with a multi-head self-attention mechanism. This mechanism processes the embedded feature matrix $\mathbf{F^{\prime}}$ by projecting it into query (Q), key (K), and value (V) spaces.

% \begin{align}
% Q &= W_Q \cdot E(D') \\
% K &= W_K \cdot E(D') \\
% V &= W_V \cdot E(D')
% \end{align}

The encoder then applies the self-attention mechanism to capture varying feature relations in the feature matrix and assigns distinct attention weights to them. Assuming $d_k$ is the dimensionality of the key vectors, the attention mechanism is formulated as:
\begin{equation}
\text{Attention}(Q, K, V) = \text{softmax}\left(\frac{QK^T}{\sqrt{d_k}}\right)V ,
\label{eq_self_attention}
\end{equation}
where $Q = W_Q \cdot \mathbf{F}^{\prime}$, $K = W_K \cdot \mathbf{F}^{\prime}$, and $V = W_V \cdot \mathbf{F}^{\prime}$, $W_Q$, $W_K$, $W_V$ are parameter matrices.

In our method, we adopt the multi-head attention mechanism, where the results of each head are concatenated and linearly transformed. Assuming $W^O$ is an output projection matrix and $\mathbf{Z}$ is the feature representation:

\begin{equation}
\text{head}_i = \text{Attention}(QW_i^Q, KW_i^K, VW_i^V) ,\\
\end{equation}
\begin{equation}
    \text{MultiHead}(Q, K, V) = \text{Concat}(\text{head}_1,..., \text{head}_h)W^O ,\\
\label{eq_multihead}
\end{equation}
\begin{equation}
    \mathbf{Z} = \text{ResNet}(\text{MultiHead}(Q, K, V)) ,
\end{equation}

\noindent where $W_i^Q$, $W_i^K$, and $W_i^V$ are weights for query, key, and value. 
Through this process, we obtain the feature representation $\mathbf{Z}$ that captures feature relationships. Specifically, $\mathbf{Z}$ is obtained by passing the input feature matrix through multiple layers of the encoder, where each layer applies self-attention and residual connection-enhanced feedforward networks.

% After processing through the multi-head attention mechanism, the output of the encoder is then aggregated to form the feature representation $\mathbf{Z}$. This is achieved by combining the outputs of all attention heads and passing them through a final transformation layer:

% \begin{equation}
% % \mathbf{Z} = \text{Transform}(\text{MultiHead}(Q, K, V))
% \mathbf{Z} = \text{Linear}(\text{MultiHead}(Q, K, V))
% \end{equation}

\noindent \textbf{Weight Decoding.} In this process, we aim to decode a weighting matrix $\mathbf{W}$ from the embedding $\mathbf{Z}$. This decoding process iteratively updates $\mathbf{W}$ until the downstream task's performance is satisfied. We initialize the $\mathbf{W}$ by setting all its elements as 1. This is to ensure all features receive equal importance at the beginning. In each decoding layer, we do cross-attention on $\mathbf{W}$ and $\mathbf{Z}$ by:
\begin{equation}
\text{CrossAttention}(Q_{W}, K_Z, V_Z) = \text{softmax}\left(\frac{Q_{W} K_Z^T}{\sqrt{d_z}}\right)V_Z ,
\label{eq_cross_attention}
\end{equation}
where $Q_w = W_Q \cdot \mathbf{W}$, $K_Z = W_K \cdot \mathbf{Z}$, and $V = W_V \cdot \mathbf{Z}$, $W_Q$, $W_K$, $W_V$ are parameter matrices.
% At the beginning of the decoding process, we introduce the initial weighting matrix $\mathbf{W}_0$, which has the same dimensions as the aligned feature $\mathbf{F^\prime}$. To initialize $\mathbf{W}_0$ and ensure that all features receive equal attention initially, we set every element in $\mathbf{W}_0$ to 1.
% Then we face the challenge of accurately extracting and integrating key information from multiple features. To tackle this challenge, we introduce a decoder that adopts a self-attention mechanism as well as a cross-attention mechanism. The self-attention mechanism allows different features to receive attention based on their importance. The cross-attention captures the relations between features. The cross-attention mechanism is denoted as:

% This method yields a weighted feature representation named $\mathbf{W}_f$. To further integrate this information and account for their interdependencies, we combine $\mathbf{W}_f$ with the feature representation $\mathbf{Z}$ from \textit{Data Encoding} with dimensionality $ d_z $. This combination is achieved through a cross-attention mechanism:

\par By adopting a cross-attention mechanism, we generate a contextual representation that captures various relationships and dependencies in the feature matrix. After several weight decoding layers, we get an updated weighting matrix $\mathbf{W}$: 
\begin{equation}
\mathbf{W} = \text{ResNet}(\text{CrossAttention}(Q_{W}, K_Z, V_Z)) .
\end{equation}

We finally use the the weighting matrix $\mathbf{W}$ to derive a weighted feature matrix $\mathbf{F}_{\text{rew}}$ by its Hadamard product with the original feature matrix $\mathbf{F}$: $\mathbf{F}_{\text{rew}} = \mathbf{W} \odot \mathbf{F}$. With this weighted feature matrix, we reorganize the feature space and make features optimized for the downstream task. $\mathbf{F}_{\text{rew}}$ is then used to substitute $\mathbf{F}$ in the downstream tasks.

\subsection{Fine-Tuning}

\par In the fine-tuning process, our primary goal is to adopt a reinforcement learning strategy to adjust the weighting matrix $\mathbf{W}$. 
% Proximal Policy Optimization (PPO) for further adjustment on weighted feature matrix $\mathbf{F}_\text{rew}$. 
This adjustment aims to reduce information redundancy of $\mathbf{F}_\text{rew}$, thereby reducing the variance during training.
% \par
% \noindent \textbf{Set up PPO Policy.} We start by defining our baseline state with the table state $D^{\prime}_0$, the Transformer's weighting matrix $W'_0$, and the initial redundancy $Rdd_0$. In the process of PPO optimization, the choice of weight distribution becomes a key concern. To tackle this challenge, we employ PPO to formulate a proposed weight distribution, denoted by $V$. This distribution is based on the table state $D^{\prime}_i$ where $i = 0, 1, \dots$. The PPO policy, denoted as $pi_i$, inputs the previous table state $D^{\prime}_{i-1}$ to capture the dynamic shifts and dependencies of the data. This strategy offers a varying probability distribution dataset based on the input to generate a recommended weighting matrix. Each weight value in the recommended weighting matrix is randomly obtained from the corresponding probability distribution. The distributions in the probability distribution dataset are distinct Gaussian distributions, represented as:
\par
\noindent \textbf{Weighting Matrix Refinement.} 
% We design a PPO network to adjust the weighting matrix $\mathbf{W}_{rew}$. In this process, the reward is redundancy, emphasizing the goal of minimizing system redundancy for optimal performance.
We begin by evaluating the redundancy, denoted as $Rdd$, using mutual information as defined by~\cite{shannon1948mathematical}. $Rdd$ is calculated as follows:
\begin{equation}
    Rdd = \frac{1}{|\mathbf{F}|^2} \sum_{\mathbf{f}_m, \mathbf{f}_n \in \mathbf{F}} I(\mathbf{f}_m,\mathbf{f}_n) .
\label{eq_Rdd}
\end{equation}
In this formula, $\mathbf{F}$ represents the feature matrix, with $\mathbf{f}_m$ and $\mathbf{f}_n$ being the $m$-th and $n$-th features, respectively. The function $I(\mathbf{f}_m,\mathbf{f}_n)$ measures the mutual information between these two features. We further define $\Delta Rdd$ to represent the change in redundancy, where $\Delta Rdd = Rdd^{\prime} - Rdd$, $Rdd^{\prime}$ is the redundancy of the feature matrix after fine-tuning. 

\par Next, we process the input weighting matrix $\mathbf{W}$ through a RL model. In this paper, we adopt a Proximal Policy Optimization (PPO) as our RL model, which comprises one \textit{actor network} and one \textit{critic network} \cite{schulman2017proximal}. While the actor network focuses on determining the actions to take, the critic evaluates how good those actions are, based on the expected rewards. In this content, an action is defined as the output of the PPO network, which is an adjusted weighting matrix $\mathbf{W}^{\prime}$. The state is the weighting matrix $\mathbf{W}$ and the reward is the change of redundancy $\Delta Rdd$.

Specifically, the actor network processes $\mathbf{W}$ to produce mean and variance values. These values are then used to form a probability distribution matrix $\mathbf{V}$, which consists of Gaussian distributions, represented as:
\begin{equation}
\mathbf{V} = (V_k^1, V_k^2, \ldots, V_k^N) ,
\label{eq_pro_matrix}
\end{equation}
\begin{equation}
    V_k^i(\mu, \sigma^2) = \frac{1}{\sqrt{2\pi\sigma^2}} e^{-\frac{(w_k^i-\mu)^2}{2\sigma^2}} .
\label{eq_pro}
\end{equation}
Here, ${V}_k^i$ indicates the weight distribution of $w_k^i$, the $i$-th element of the $k$-th feature, with $\mu$ and $\sigma^2$ being the mean and variance, respectively. Here we form $\mathbf{W}^{\prime}$ with each element ${w^{\prime}}_k^i$ sampled from the probability distribution $V_k^i$.

\noindent \textbf{Actor Network Update.} After refining the weighting matrix, we update the feature matrix as $\mathbf{F}_\text{rew}^{\prime} = \mathbf{W}^{\prime} \odot \mathbf{F}$, and subsequently calculate the information redundancy $Rdd^{\prime}$ of $\mathbf{F}_\text{rew}^{\prime}$. Based on the observed change of redundancy $\Delta Rdd$, we adjust the mean and variance within the probability distribution, following the equations:
\begin{equation}
    {\theta}^{\prime} \leftarrow {\theta} + \alpha \cdot \Delta Rdd \cdot \nabla J(\theta) ,
\label{eq_actor_parameter}
\end{equation}
\begin{equation}
    \nabla J(\theta) = \frac{1}{n} \sum_{i=1}^{n} (\nabla \log \pi_{\theta} (a_i | s_i)) R_i ,
\end{equation}
where $ \theta $ is the parameter of the actor network, $ \alpha $ is the learning rate, $ J(\theta) $ is the objective function to maximize, $ \nabla J(\theta) $ is the gradient of the objective function with respect to the mean and variance, $n$ is the number of state-action pairs in the training, $\pi_\theta(a|s)$ is the policy, and $R_i$ is the reward of state-action pair $(s_i, a_i)$. To ensure a stable fine-tuning process, we implement a clipping mechanism for the updated means. Specifically, we adopt each mean ${\mu_i}$ using the formula: ${\mu_i} = \text{clip}({\mu}_i, {w_i}+\epsilon, {w_i}-\epsilon)$. This clipping process is crucial for as it prevents excessive deviations from the current weight $w_i$, thereby maintaining the stability and reducing the variance during downstream training.

\noindent \textbf{Critic Network Update.} After the update of actor network, we continue to adjust the critic network with the reward. We design the critic network to provide an estimate of the advantage function $A(s,a)$. This advantage function represents the expected future advantages under the state $s$ and action $a$. 
We design the function to change the policy gradually based on the current state, so that the policy after adjustment $\pi'_\theta(a|s)$ is not too biased from the previous policy $\pi_\theta(a|s)$. 
We adopt the clipping mechanism with a parameter $\epsilon$ as well as the advantage function $A(s, a)$ in the loss function:

\begin{align}
L(\theta) &= \mathbb{E}_{a,s} \Bigg[ \min \left( \frac{\pi_\theta(a|s)}{\pi'_\theta(a|s)} A(s,a), \right. \notag \\
& \quad \left. \text{clip}\left(\frac{\pi_\theta(a|s)}{\pi'_\theta(a|s)}, 1-\epsilon, 1+\epsilon\right) A(s,a) \right] .
\label{eq_ppo_loss}
\end{align}
By minimizing $L(\theta)$, we continuously optimize the feature matrix to obtain stable and enhanced performance.
% of downstream tasks.

% \begin{algorithm}
% \caption{weighting Matrix Adjustment Process}
% \begin{algorithmic}
% \REQUIRE weighting matrix $\mathbf{W}$, feature matrix $\mathbf{F}$,iteration $K$

% \FOR{$k = 1,2,...,K$}
%     \STATE Multiply $\mathbf{W}$ with data to get weighted feature matrix $\mathbf{F}_{\text{rew}}$
%     \STATE Pass $\mathbf{W}$ through an actor network to obtain mean $\mu$ and variance $\sigma$
%     \STATE Sample adjusted weighting matrix $\mathbf{W}^{\prime}$ from Gaussian distribution $\mathcal{N}(\mu, \sigma)$
%     \STATE Multiply $\mathbf{W^{\prime}}$ with $\mathbf{F}$ to get $\mathbf{F^{\prime}}_{\text{rew}}$
%     \STATE Set $\mathbf{F} = \mathbf{F^\prime}_{\text{rew}}$
% \ENDFOR
% \RETURN PPO weighting matrix $\mathbf{W^{\prime}}$
%     \end{algorithmic}
% \end{algorithm}

\begin{algorithm}[h]

\DontPrintSemicolon
  
  \KwInput{dataset $\mathcal{D} = \{\mathbf{F}, \mathbf{y}\}$}
  \KwOutput{weighted feature matrix $\mathbf{F}^{\prime}_{\text{rew}}$}
\For{ iteration = $0,1,2,\ldots, I$}
{
  Convert $\mathbf{F}$ into $\mathbf{F^{\prime}}$ by Eq.\ref{eq_discrete} and Eq.\ref{eq_contious}
  
  % Let $\mathbf{Z} \leftarrow \mathbf{F^{\prime}}$
  
  \For{each encoder}    
        { 
        	$Q \leftarrow W_Q \cdot \mathbf{F}^{\prime}$, $K \leftarrow W_K \cdot \mathbf{F}^{\prime}$, and $V \leftarrow W_V \cdot \mathbf{F}^{\prime}$

          Compute $\text{MultiHead}(Q, K, V)$ by Eq.\ref{eq_self_attention} and Eq.\ref{eq_multihead}

          $\mathbf{Z} \leftarrow \text{ResNet}(\text{MultiHead}(Q, K, V))$
          
        }
  % Let $\mathbf{W} \leftarrow \mathbf{W}_0$  
  
  \For{each decoder}    
        { 
        	$Q_w \leftarrow W_Q \cdot \mathbf{W}$, $K_Z \leftarrow W_K \cdot \mathbf{Z}$, and $V \leftarrow W_V \cdot \mathbf{Z}$

          Compute $\text{CrossAttention}(Q_{W}, K_Z, V_Z)$ by Eq.\ref{eq_cross_attention}

          $\mathbf{W} \leftarrow \text{ResNet}(\text{CrossAttention}(Q_{W}, K_Z, V_Z))$
          
        } 
  $\mathbf{F}_{\text{rew}} \leftarrow \mathbf{W} \odot \mathbf{F}$
  
  \For{each state-action pairs}    
        { 
          Get $(\mu,\sigma)$ by process $\mathbf{W}$ through PPO network
  
          Estimate $\mathbf{V}$ by $(\mu,\sigma)$ by Eq.\ref{eq_pro_matrix} and Eq.\ref{eq_pro}
        
          Sample $\mathbf{W}^{\prime}$ from $\mathbf{V}$
        
          $\mathbf{F^{\prime}}_{\text{rew}} \leftarrow \mathbf{W^{\prime}} \odot \mathbf{F}$

          Compute $Rdd$ for $\mathbf{F}_{\text{rew}}$, $Rdd^{\prime}$ for $\mathbf{F^{\prime}}_{\text{rew}}$ by Eq.\ref{eq_Rdd}

          $\Delta Rdd \leftarrow Rdd^{\prime} - Rdd$
          % , $Rdd^{\prime}$
        }
    Update actor network parameter by Eq.\ref{eq_actor_parameter} 
    
    Update critic network parameter by Eq.\ref{eq_ppo_loss}

    Pretrain a predictive model $\mathcal{M}$ with $\mathcal{D}$

   Get $\mathbf{\hat{y}}$ by perform $\mathcal{M}$ on $\mathbf{F}^{\prime}_{\text{rew}}$, and compute cross-entropy loss between $\mathbf{\hat{y}}$ and $\mathbf{y}$.

   Backpropagate the loss to update parameters in the $\textit{encoders}$ and $\textit{decoders}$.
}
\caption{Training of TFWT}
\label{Alg_TFWT}
\end{algorithm}

\subsection{Training of TFWT} As Algorithm \ref{Alg_TFWT} shows, we first align original features by Eq.\ref{eq_discrete} and Eq.\ref{eq_contious}. Then, we encode the aligned feature matrix $\mathbf{F^{\prime}}$ into an embedding $\mathbf{Z}$ and decode it into a weighting matrix $\mathbf{W}$. This encoding-decoding process is accomplished by a designated Transformer. In this way, we get a weighted feature matrix $\mathbf{F}_{\text{rew}}$. To further fine-tune $\mathbf{W}$, we adopt PPO as a reinforcement learning model to reduce the redundancy of $\mathbf{F}_{\text{rew}}$. The fine-tuned $\mathbf{W}^{\prime}$ is sampled from the actor network of PPO. The PPO networks are trained by the interaction data from the sampling process. The cross-entropy loss derived from the downstream machine learning model is used to update the parameters of the encoders and decoders in the Transformer.

\section{Experiments}
\par In this section, we present three experiments that validate the strength of our method. First, we demonstrate that our method significantly enhances the performance on various downstream tasks without fine-tuning. Second, we demonstrate the advantages of our TFWT method over the baseline methods. Finally, we demonstrate the effectiveness of fine-tuning comparing with non-fine-tuning version of TFWT in reducing the variance performance metrics. Overall, the results consistently demonstrate the superior performance of our method.

\begin{table*}
  \centering
  % \vspace{-2mm}
  \resizebox{1.02\textwidth}{!}{% Resize table to 80% of text width
    \begin{tabular}{|c|l|cccc|cccc|cccc|cccc|cccc|}
    \hline
    \multirow{3}[2]{*}{\textbf{Metrics}} & \multicolumn{1}{c|}{\multirow{3}[2]{*}{\textbf{Model}}} & \multicolumn{4}{c|}{\multirow{3}[2]{*}{\textbf{RF}}} & \multicolumn{4}{c|}{\multirow{3}[2]{*}{\textbf{LR}}} & \multicolumn{4}{c|}{\multirow{3}[2]{*}{\textbf{NB}}} & \multicolumn{4}{c|}{\multirow{3}[2]{*}{\textbf{KNN}}} & \multicolumn{4}{c|}{\multirow{3}[2]{*}{\textbf{MLP}}} \\
          &       & \multicolumn{4}{c|}{}         & \multicolumn{4}{c|}{}         & \multicolumn{4}{c|}{}         & \multicolumn{4}{c|}{}         & \multicolumn{4}{c|}{} \\
          &       & \multicolumn{4}{c|}{}         & \multicolumn{4}{c|}{}         & \multicolumn{4}{c|}{}         & \multicolumn{4}{c|}{}         & \multicolumn{4}{c|}{} \\
    \hline
    \multirow{7}[14]{*}[4ex]{\textbf{Acc}} &  \rule{0pt}{10pt} & \multicolumn{1}{c}{AM} & \multicolumn{1}{c}{OS} & \multicolumn{1}{c}{MA} & \multicolumn{1}{c|}{SD} & \multicolumn{1}{c}{AM} & \multicolumn{1}{c}{OS} & \multicolumn{1}{c}{MA} & \multicolumn{1}{c|}{SD} & \multicolumn{1}{c}{AM} & \multicolumn{1}{c}{OS} & \multicolumn{1}{c}{MA} & \multicolumn{1}{c|}{SD} & \multicolumn{1}{c}{AM} & \multicolumn{1}{c}{OS} & \multicolumn{1}{c}{MA} & \multicolumn{1}{c|}{SD} & \multicolumn{1}{c}{AM} & \multicolumn{1}{c}{OS} & \multicolumn{1}{c}{MA} & \multicolumn{1}{c|}{SD} \\
\cline{2-22}          & Raw \rule{0pt}{10pt}   & \multicolumn{1}{c}{0.620 } & 0.850  & 0.812  & 0.702  & 0.660  & 0.873  & 0.787  & 0.724  & 0.600  & 0.780  & 0.731  & 0.679  & 0.567  & 0.843  & 0.808  & 0.674  & 0.687  & 0.868  & 0.812  & 0.716  \\
\cline{2-2}          & USP \rule{0pt}{10pt}   & 0.613  & 0.863  & 0.817  & 0.682  & 0.640  & 0.869  & 0.751  & 0.710  & 0.587  & 0.766  & \underline{0.744}  & 0.692  & 0.560  & 0.858  & 0.815  & 0.662  & 0.653  & 0.869  & 0.819  & 0.720  \\
\cline{2-2}          & Lasso \rule{0pt}{10pt} & 0.627  & 0.860  & 0.822  & 0.707  & \underline{0.680}  & 0.860  & \underline{0.799}  & 0.724  & 0.607  & 0.802  & 0.738  & 0.669  & 0.533  & 0.852  & 0.816  & \underline{0.679}  & 0.693  & 0.878  & 0.821  & 0.719  \\
\cline{2-2}          & WB \rule{0pt}{10pt}  & 0.613  & 0.868  & 0.823  & 0.690  & 0.667  & 0.861  & 0.790  & 0.710  & \underline{0.633}  & 0.786  & 0.742  & \underline{0.692}  & 0.567  & 0.857  & 0.822  & 0.664  & \underline{0.707}  & 0.869  & 0.821  & 0.722  \\
\cline{2-2}          & TabT \rule{0pt}{10pt}  & \underline{0.620}  & \underline{0.882}  & \underline{0.851}  & \underline{0.708}  & 0.660  & \underline{0.879}  & 0.797  & \underline{0.725}  & 0.600  & \underline{0.817}  & 0.741  & 0.688  & \underline{0.567}  & \underline{0.869}  & \underline{0.822}  & 0.678  & 0.687  & \underline{0.878}  & \underline{0.851}  & \underline{0.723}  \\
\cline{2-2}          & \textbf{TFWT} \rule{0pt}{10pt}  & \textbf{0.640 } & \textbf{0.895 } & \textbf{0.860 } & \textbf{0.733 } & \textbf{0.713 } & \textbf{0.903 } & \textbf{0.805 } & \textbf{0.727 } & \textbf{0.627 } & \textbf{0.829 } & \textbf{0.752 } & \textbf{0.694 } & \textbf{0.587 } & \textbf{0.884 } & \textbf{0.833 } & \textbf{0.685 } & \textbf{0.727 } & \textbf{0.894 } & \textbf{0.874 } & \textbf{0.739 } \\
    \hline
    \multirow{7}[14]{*}[4ex]{\textbf{Prec}} &  \rule{0pt}{10pt} & \multicolumn{1}{c}{AM} & \multicolumn{1}{c}{OS} & \multicolumn{1}{c}{MA} & \multicolumn{1}{c|}{SD} & \multicolumn{1}{c}{AM} & \multicolumn{1}{c}{OS} & \multicolumn{1}{c}{MA} & \multicolumn{1}{c|}{SD} & \multicolumn{1}{c}{AM} & \multicolumn{1}{c}{OS} & \multicolumn{1}{c}{MA} & \multicolumn{1}{c|}{SD} & \multicolumn{1}{c}{AM} & \multicolumn{1}{c}{OS} & \multicolumn{1}{c}{MA} & \multicolumn{1}{c|}{SD} & \multicolumn{1}{c}{AM} & \multicolumn{1}{c}{OS} & \multicolumn{1}{c}{MA} & \multicolumn{1}{c|}{SD} \\
\cline{2-22}          & Raw \rule{0pt}{10pt}  & 0.622  & 0.727  & 0.827  & 0.706  & 0.657  & 0.789  & 0.777  & 0.701  & 0.613  & 0.659  & 0.739  & 0.681  & 0.537  & 0.707  & 0.838  & 0.656  & 0.687  & 0.788  & 0.822  & 0.716  \\
\cline{2-2}          & USP \rule{0pt}{10pt}   & \underline{0.629}  & 0.714  & 0.801  & 0.706  & 0.634  & 0.761  & 0.754  & 0.708  & 0.583  & 0.649  & 0.716  & 0.694  & \textbf{0.633}  & 0.786  & 0.828  & 0.683  & 0.664  & 0.766  & 0.810  & \underline{0.736}  \\
\cline{2-2}          & Lasso \rule{0pt}{10pt} & 0.624  & 0.747  & 0.816  & 0.694  & \underline{0.684}  & 0.771  & \underline{0.789}  & 0.724  & 0.607  & 0.680  & 0.732  & 0.670  & 0.544  & \textbf{0.851 } & 0.831  & 0.679  & 0.689  & 0.806  & 0.820  & 0.725  \\
\cline{2-2}          & WB \rule{0pt}{10pt}  & 0.614  & 0.765  & \textbf{0.876 } & \underline{0.753}  & 0.670  & 0.775  & 0.788  & 0.710  & \underline{0.619}  & 0.662  & 0.728  & \textbf{0.699 } & 0.557  & 0.790  & \underline{0.840}  & 0.660  & \underline{0.707}  & 0.793  & 0.835  & 0.722  \\
\cline{2-2}          & TabT \rule{0pt}{10pt} & 0.622  & \underline{0.788}  & \underline{0.860}  & 0.710  & 0.657  & \underline{0.799}  & 0.786  & \underline{0.725}  & 0.613  & \underline{0.687}  & \underline{0.742}  & 0.688  & 0.537 & 0.787  & 0.821  & \underline{0.688}  & 0.687  & \underline{0.812}  & \underline{0.840}  & 0.731  \\
\cline{2-2}          & \textbf{TFWT}  \rule{0pt}{10pt} & \textbf{0.637 } & \textbf{0.856 } & 0.842  & \textbf{0.742 } & \textbf{0.710}  & \textbf{0.803 } & \textbf{0.789 } & \textbf{0.727 } & \textbf{0.627 } & \textbf{0.694 } & \textbf{0.769 } & \underline{0.696}  & \underline{0.557}  & \underline{0.801}  & \textbf{0.845 } & \textbf{0.707 } & \textbf{0.730 } & \textbf{0.825 } & \textbf{0.871 } & \textbf{0.739 } \\
    \hline
    \multirow{7}[14]{*}[4ex]{\textbf{Rec}} &  \rule{0pt}{10pt} & \multicolumn{1}{c}{AM} & \multicolumn{1}{c}{OS} & \multicolumn{1}{c}{MA} & \multicolumn{1}{c|}{SD} & \multicolumn{1}{c}{AM} & \multicolumn{1}{c}{OS} & \multicolumn{1}{c}{MA} & \multicolumn{1}{c|}{SD} & \multicolumn{1}{c}{AM} & \multicolumn{1}{c}{OS} & \multicolumn{1}{c}{MA} & \multicolumn{1}{c|}{SD} & \multicolumn{1}{c}{AM} & \multicolumn{1}{c}{OS} & \multicolumn{1}{c}{MA} & \multicolumn{1}{c|}{SD} & \multicolumn{1}{c}{AM} & \multicolumn{1}{c}{OS} & \multicolumn{1}{c}{MA} & \multicolumn{1}{c|}{SD} \\
\cline{2-22}          & Raw \rule{0pt}{10pt}   & 0.622  & 0.722  & 0.761  & 0.702  & 0.655  & 0.658  & 0.741  & 0.701  & 0.602  & 0.729  & 0.661  & 0.680  & 0.705  & 0.631  & 0.749  & 0.636  & 0.687  & 0.662  & 0.772  & 0.716  \\
\cline{2-2}          & USP \rule{0pt}{10pt}   & 0.582  & 0.714  & 0.818  & 0.616  & 0.633  & \textbf{0.781}  & \textbf{0.779 } & 0.709  & 0.579  & 0.721  & \underline{0.666}  & 0.681  & 0.524  & 0.609  & 0.761  & 0.659  & 0.651  & 0.668  & 0.793  & 0.719  \\
\cline{2-2}          & Lasso \rule{0pt}{10pt} & 0.622  & 0.700  & 0.789  & \textbf{0.736 } & \underline{0.678}  & 0.633  & 0.748  & 0.724  & 0.603  & \underline{0.771}  & 0.661  & 0.659  & 0.510  & \textbf{0.812 } & 0.758  & 0.679  & 0.690  & \underline{0.700}  & 0.780  & 0.718  \\
\cline{2-2}          & WB \rule{0pt}{10pt}  & 0.614  & \underline{0.715}  & 0.748  & 0.562  & 0.667  & 0.638  & 0.747  & 0.705  & \underline{0.616}  & 0.727  & 0.649  & 0.672  & 0.546  & 0.602  & 0.754  & 0.670  & \underline{0.710 } & 0.665  & 0.769  & 0.722  \\
\cline{2-2}          & TabT \rule{0pt}{10pt}  & \underline{0.622}  & \textbf{0.715}  & \underline{0.821}  & 0.708  & 0.655  & 0.655  & 0.755  & 0.725  & 0.602  & \textbf{0.777 } & \textbf{0.668 } & \underline{0.688}  & \underline{0.705}  & \underline{0.668}  & \textbf{0.785 } & 0.678  & 0.687  & 0.635  & \underline{0.839}  & \underline{0.723}  \\
\cline{2-2}          & \textbf{TFWT} \rule{0pt}{10pt}  & \textbf{0.641 } & 0.688  & \textbf{0.847 } & \underline{0.732}  & \textbf{0.710 } & \underline{0.741}  & \underline{0.761}  & \textbf{0.727 } & \textbf{0.634 } & 0.750  & 0.658  & \textbf{0.693 } & \textbf{0.782 } & 0.626  & \underline{0.774}  & \textbf{0.685 } & \textbf{0.730}  & \textbf{0.701 } & \textbf{0.840 } & \textbf{0.739 } \\
    \hline
    \multirow{7}[14]{*}[4ex]{\textbf{F1}} &  \rule{0pt}{10pt} & \multicolumn{1}{c}{AM} & \multicolumn{1}{c}{OS} & \multicolumn{1}{c}{MA} & \multicolumn{1}{c|}{SD} & \multicolumn{1}{c}{AM} & \multicolumn{1}{c}{OS} & \multicolumn{1}{c}{MA} & \multicolumn{1}{c|}{SD} & \multicolumn{1}{c}{AM} & \multicolumn{1}{c}{OS} & \multicolumn{1}{c}{MA} & \multicolumn{1}{c|}{SD} & \multicolumn{1}{c}{AM} & \multicolumn{1}{c}{OS} & \multicolumn{1}{c}{MA} & \multicolumn{1}{c|}{SD} & \multicolumn{1}{c}{AM} & \multicolumn{1}{c}{OS} & \multicolumn{1}{c}{MA} & \multicolumn{1}{c|}{SD} \\
\cline{2-22}          & Raw \rule{0pt}{10pt}  & 0.620  & 0.725  & 0.777  & 0.701  & 0.656  & 0.694  & 0.752  & 0.701  & 0.591  & 0.676  & 0.666  & 0.679  & 0.432  & 0.654  & 0.766  & 0.624  & 0.687  & 0.697  & 0.785  & 0.716  \\
\cline{2-2}          & USP \rule{0pt}{10pt}   & 0.556  & 0.734  & 0.807  & 0.658  & 0.633  & \textbf{0.770}  & 0.746  & 0.709  & 0.578  & 0.664  & 0.677  & 0.687  & 0.418  & 0.637  & 0.779  & 0.650  & 0.646  & 0.699  & 0.800  & 0.714  \\
\cline{2-2}          & Lasso \rule{0pt}{10pt} & \underline{0.622}  & 0.719  & 0.798  & \underline{0.714}  & \underline{0.676}  & 0.665  & 0.761  & 0.724  & 0.602  & 0.703  & \underline{0.669}  & 0.664  & 0.403  & \textbf{0.826}  & 0.776  & \textbf{0.679}  & 0.690  & \underline{0.735}  & 0.793  & 0.717  \\
\cline{2-2}          & WB \rule{0pt}{10pt}  & 0.613  & \underline{0.736}  & 0.772  & 0.644  & 0.666  & 0.671  & 0.758  & 0.708  & \underline{0.617}  & 0.680  & 0.658  & 0.685  & \textbf{0.532}  & 0.629  & 0.776  & 0.665  & \underline{0.705}  & 0.700  & 0.787  & \underline{0.722}  \\
\cline{2-2}          & TabT \rule{0pt}{10pt}  & 0.620  & \textbf{0.743}  & \underline{0.834}  & 0.707  & 0.656 & 0.693  & \underline{0.765}  & \underline{0.725}  & 0.591  & \underline{0.712}  & \textbf{0.677 } & \underline{0.688}  & 0.432  & \underline{0.703}  & \textbf{0.797}  & 0.674  & 0.687  & 0.673  & \underline{0.839}  & 0.720  \\
\cline{2-2}          & \textbf{TFWT} \rule{0pt}{10pt}  & \textbf{0.636 } & 0.735  & \textbf{0.844 } & \textbf{0.730 } & \textbf{0.710}  & \underline{0.767 } & \textbf{0.771 } & \textbf{0.727 } & \textbf{0.621 } & \textbf{0.715 } & 0.667  & \textbf{0.692 } & \underline{0.463 } & 0.664  & \underline{0.794 } & \underline{0.676 } & \textbf{0.730 } & \textbf{0.742 } & \textbf{0.852 } & \textbf{0.728 } \\
    \hline
    \end{tabular}%
  }
  \caption{Overall performance on downstream tasks. The best results are highlighted in \textbf{bold}, and the runner-up results are highlighted in \underline{underline}. (Higher values indicate better performance.)}
  \vspace{-3mm} 
  \label{tab:overallperformance}%
\end{table*}

\begin{table}
\small
\centering
% \vspace{-1mm}
\setlength{\tabcolsep}{8pt}
\begin{tabular}{l|lll}
    \toprule
    Datasets & Samples & Features & Class \\
    \midrule
    AM    & 1,500  & 10,000     & 2 \\
    OS    & 12,330 & 17    & 2 \\
    MA    & 19,020 & 10    & 2 \\
    SD    & 991,346 & 23    & 2 \\
    \bottomrule
    \end{tabular}%
\caption{Datasets Description.}
% \vspace{-3mm}
\label{tab:plain}
\end{table}

\subsection{Experimental Settings}
\noindent \textbf{Datasets.} We evaluate the proposed method with four real-world datasets:
\begin{itemize}
    \item \textbf{Amazon Commerce Reviews Set (AM)}~\cite{misc_amazon_commerce_reviews_set_215} from UCI consists of customer reviews from the Amazon Commerce website. Its purpose is to classify the identities of authors of reviews by analyzing textual patterns. We have randomly divided its $50$ labels into two groups, each containing $25$ labels, transforming it into a balanced binary classification task.
    \item \textbf{Online Shoppers Purchasing Intention Dataset (OS)}~\cite{misc_online_shoppers_purchasing_intention_dataset_468} from UCI features multivariate data types including integer and real values. Its purpose is to classify shoppers' purchasing intentions and predict purchases.

    \item \textbf{MAGIC Gamma Telescope Dataset (MA)}~\cite{misc_magic_gamma_telescope_159} from UCI reflects the simulation of high energy gamma particles registration in a gamma telescope. Its purpose is to classify the primary gammas from cosmic rays.
    \item \textbf{Smoking and Drinking Dataset with body signal (SD)}~\cite{kaggle_smoking_drinking} from Kaggle is collected from National Health Insurance Service in Korea. Its purpose is to classify ``smoker" or ``drinker".

\end{itemize}

\noindent \textbf{Downstream Tasks.} We apply the proposed model across a diverse array of classification tasks, including \textit{Random Forests (RF)}, \textit{Logistic Regression (LR)}, \textit{Naive Bayes (NB)}, \textit{K-Nearest Neighbor (KNN)} and \textit{Multilayer Perceptrons (MLP)}. We compare the performance outcomes in these tasks both with and without our method.

\noindent \textbf{Baseline Models.} To demonstrate the effectiveness of our method, We compare our TFWT method with four established baseline techniques, where the Least Absolute Shrinkage and Selection Operator and TabTransformer are used for feature preprocessing, and Weighted Bootstrapping and Undersampling handle sample weight preprocessing.

\begin{itemize}
    \item \textit{Undersampling (USP)} reduces the majority class in a dataset to balance with the minority class, creating an even dataset. This method minimizes majority class bias in training. We set the undersampling ratio based on category frequency in our experiments.
    
    \item \textit{Least Absolute Shrinkage and Selection Operator (LASSO)} is a technique for feature selection and regularization, enhancing model accuracy and interpretability. It introduces a penalty proportional to the absolute values of coefficients, encouraging sparsity by driving some to zero. This process effectively selects crucial features, simplifying the model and reducing data dimensionality.

    \item \textit{Weighted Bootstrapping (WB)}~\cite{barbe1995weighted} is a resampling technique assigning weights to each dataset instance, influencing their selection in the resampled dataset. It's particularly useful for balancing underrepresented classes. In our experiments, weights are determined by class frequency.
    
    \item \textit{TabTransformer (TabT)}~\cite{huang2020tabtransformer} is a method designed for tabular data, inspired by Transformer technology from natural language processing. It specializes in transforming categorical features into embeddings, capturing complex relationships within the data. This approach enhances the performance of tabular data in downstream tasks. In our experiments, TabTransformer processes categorical features to create enriched representations, which are then integrated into our model.

\end{itemize}
\noindent \textbf{Metrics.} To evaluate our proposed method, we use the following metrics: \textit{Overall Accuracy (Acc)} measures the proportion of true results (both true positives and true negatives) in the total dataset. \textit{Precision (Prec)} reflects the ratio of true positive predictions to all positive predictions for each class. \textit{Recall (Rec)}, also known as sensitivity, reflects the ratio of true positive predictions to all actual positives for each class. \textit{F-Measure (F1)} is the harmonic mean of precision and recall, providing a single score that balances both metrics. 

\noindent \textbf{Implementation Details.} We implemented TFWT using PyTorch and Scikit-learn. The models were trained on NVIDIA A100. For each dataset, we randomly selected between $60\%$ and $80\%$ as training data. We initialized the hyperparameters for the baseline models following the guidelines in the corresponding papers, and carefully adjusted them to ensure optimal performance. The initial learning rate was set between $10^{-3}$ and $10^{-5}$. For model regularization, the dropout rate was fixed at $0.2$.

\begin{figure}[t!]

  \centering
  \includegraphics[width=1\linewidth]{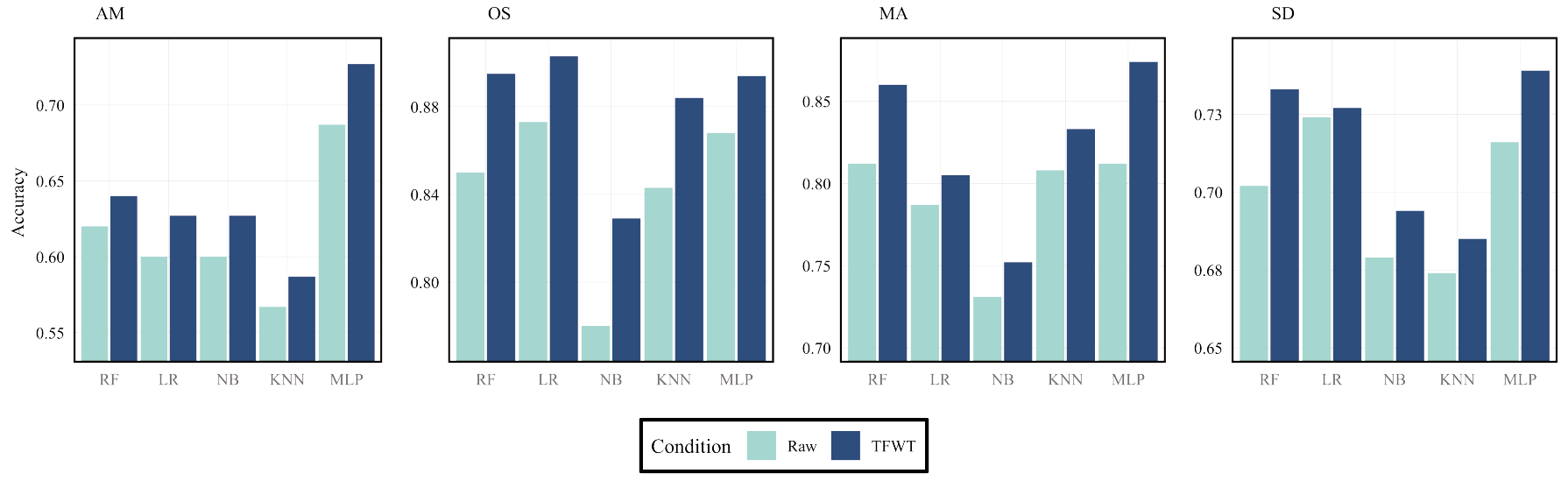}
  % \vspace{-3mm}
  \caption{ Accuracy Improvement Comparison.}
  \label{fig:Accuracy Improvement Comparison}
  % \vspace{-3mm}
\end{figure}
\subsection{Experimental Results}
% \textbf{Overall Performance.} Table~\ref{tab:overallperformance} demonstrates that TFWT consistently outperforms baseline methods in terms of all metrics across different datasets. For instance, considering MLP in Fig~\ref{fig:exp-3-2}, TFWT achieves an accuracy ranging from 117\% to 123\% compared to models without weight adjustment, and it manifests a 14.6\% to 39.8\% improvement over the most competitive weight adjustment methods. Using optimization with PPO, accuracy escalates to between 119\% and 127\%, marking an additional 6\% to 12\% enhancement over using TFWT alone. This underscores the efficacy of our TFWT-PPO approach. Our method persistently outperforms TabTransformer, which also integrates the Transformer structure, due to TFWT's weight adjustment in line with downstream task requirements and its application of an apt encoding scheme for all features.
\textbf{Overall Performance.} Table~\ref{tab:overallperformance} illustrates that our TFWT method consistently surpasses baseline methods across a variety of metrics and datasets. For example, when focusing on MLP, TFWT attains an accuracy improvement ranging from $17\%$ to $23\%$ compared to raw downstream tasks, and from $14.6\%$ to $39.8\%$ improvement over the most competitive feature weight adjustment methods. Furthermore, when applying fine-tuning method, TFWT sees an additional accuracy increase from $19\%$ to $27\%$, and a variance decline from $5\%$ to $18\%$. Notably, our method also consistently outperforms the TabTransformer model, which also incorporates the Transformer for feature adjustment.
% With fine-tuning, accuracy escalates to between 119\% and 127\%, marking an additional 6\% to 12\% enhancement over non-fine-tuned part of our method. Our method persistently outperforms TabTransformer, which also integrates the Transformer structure for feature adjustment.

% \par \textbf{Robustness Check.} To verify that predictive accuracy depends not only on feature weighting but also on the type of downstream tasks, we apply our method to a variety of predictors. This is to check if the feature subset we identified remains stable across different predictors and consistently outperforms other baseline methods. This process helps us evaluate the robustness of our method.
% \par For this experiment, we utilized various predictors including (i) Logistic Regression (LR);(ii) Naive Bayes (NB); (iii) K-Nearest Neighbors (KNN); and (iv) Multi-Layer Perceptron (MLP). Figure~\ref{fig:Accuracy Improvement Comparison} and ~\ref{fig:mlp-nb-acc} demonstrate that our TFWT method surpasses baseline methods across nearly all predictors. 

% \noindent \textbf{Comparison with Raw Downstream Tasks.} We assess the impact of combining the TFWT pretrained model with various downstream tasks on task performance. 
% Figure~\ref{fig:Accuracy Improvement Comparison} and Table~\ref{tab:overallperformance} present the comparative results, illustrating that TFWT consistently improves performance across all four metrics on four datasets, especially with MLP. The results confirm that the TFWT pretrained model indeed enhances the learning capabilities of downstream tasks from multiple perspectives.

\noindent \textbf{Enhancement over Raw Downstream Tasks.} Our evaluation focuses on the improvement that TFWT method brings to various downstream tasks in terms of performance. To ensure a robust and reliable comparison, we execute each model configuration five times and calculate the average metrics. The comparative results, showcased in Figure~\ref{fig:Accuracy Improvement Comparison} and Table~\ref{tab:overallperformance}, clearly demonstrate that TFWT consistently boosts performance across all four metrics in four datasets, particularly when applied to MLP. The significant improvement incorporated in the TFWT method enhances the performance of downstream tasks from multiple dimensions.
% These diverse comparisons highlight the adaptability and effectiveness of the TFWT method in enhancing downstream task performance.

% \begin{table}[H]
% \small
% \centering
% \scalebox{0.93}{
% \setlength{\tabcolsep}{3pt}
% \begin{tabular}
% {@{}lc|ccccc@{}}
% \toprule
% \textbf{Dataset} & \textbf{Metric} & \textbf{RF} & \textbf{LR} & \textbf{NB} & \textbf{KNN} & \textbf{MLP} \\
% \midrule
% \multirow{4}{*}{BM} & Acc & 0.902 & 0.886 & 0.817 & 0.887 & 0.835 \\
%  & Rec & 0.690 & 0.600 & 0.670 & 0.660 & 0.490 \\
%  & Prec & 0.790 & 0.750 & 0.630 & 0.740 & 0.480 \\
%  & F1 & 0.720 & 0.630 & 0.640 & 0.680 & 0.480 \\
% \midrule
% \multirow{4}{*}{CI} & Acc & 0.852 & 0.823 & 0.800 & 0.820 & 0.762 \\
%  & Rec & 0.760 & 0.690 & 0.630 & 0.730 & 0.500 \\
%  & Prec & 0.810 & 0.780 & 0.750 & 0.750 & 0.880 \\
%  & F1 & 0.780 & 0.720 & 0.650 & 0.740 & 0.430 \\
% \midrule 
% \multirow{4}{*}{DHI} & Acc & 0.860 & 0.866 & 0.774 & 0.838 & 0.853 \\
%  & Rec & 0.580 & 0.570 & 0.690 & 0.590 & 0.520 \\
%  & Prec & 0.680 & 0.710 & 0.620 & 0.630 & 0.600 \\
%  & F1 & 0.600 & 0.590 & 0.640 & 0.610 & 0.520
%  \\
% \midrule
% \multirow{4}{*}{SD} & Acc & 0.733 & 0.724 & 0.693 & 0.673 & 0.677 \\
%  & Rec & 0.730 & 0.720 & 0.690 & 0.670 & 0.680 \\
%  & Prec & 0.730 & 0.720 & 0.690 & 0.670 & 0.680 \\
%  & F1 & 0.730 & 0.720 & 0.690 & 0.670 & 0.680
%  \\
% \bottomrule
% \end{tabular}
% }
% \caption{Performance of Downstream Tasks on Various Datasets}
% \label{tab:performance-metrics}
% \end{table}

\begin{figure}[!t]
      \centering
  \includegraphics[width=1\linewidth]{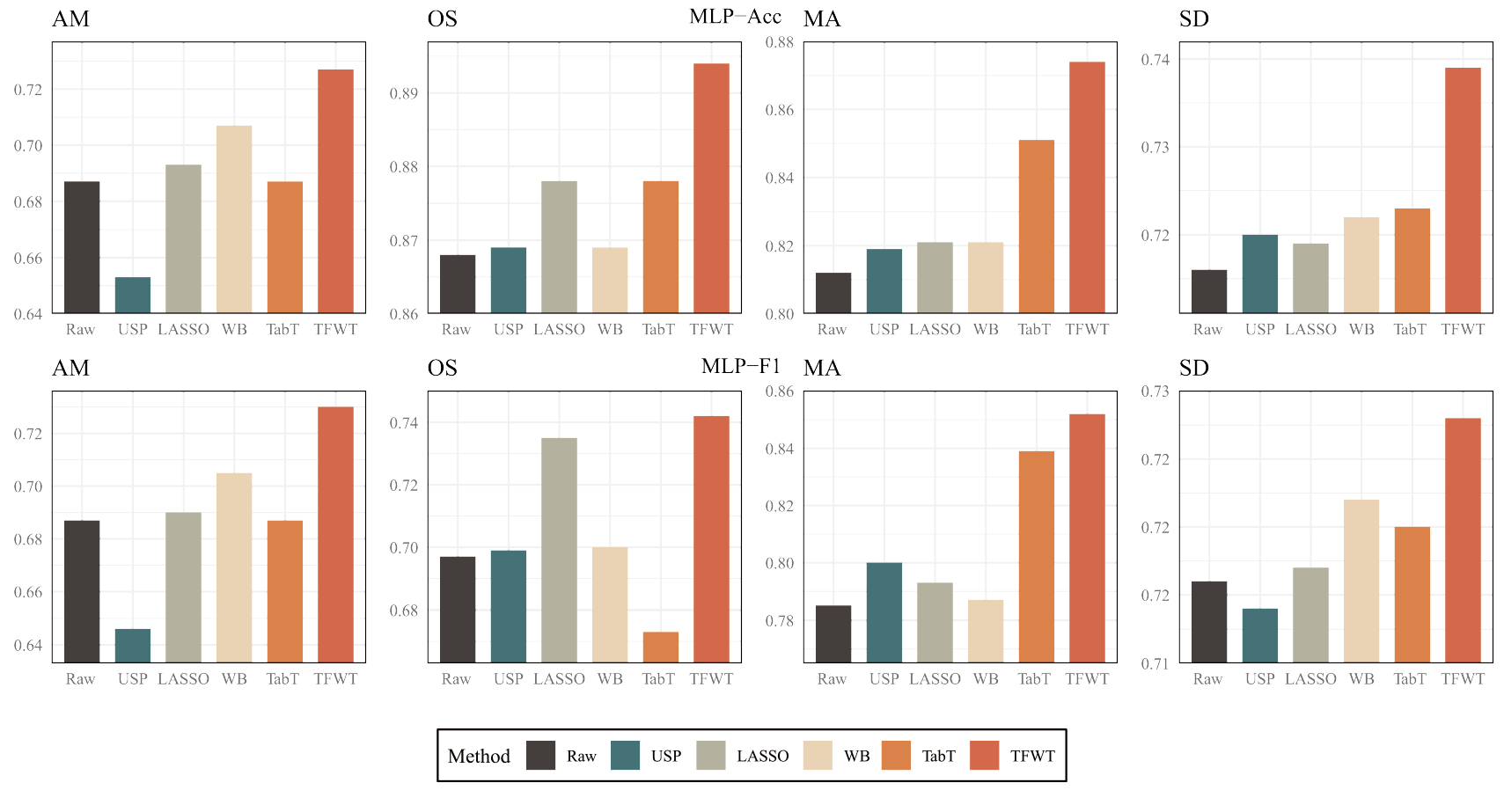}
  % \vspace{-3mm}
  \caption{Comparison on MLP (Accuracy and F1).}
  \label{fig:mlp-acc}
  % \vspace{-3mm}  
\end{figure}

% \begin{figure}[h]
%   \centering
%   \includegraphics[width=0.8\linewidth]{figure/NB-Acc.eps}
%   \caption{Comparison on NB (Accuracy)}
%   \label{fig:nb-acc}
% \end{figure}

\begin{figure}[!t]
  \centering
  % \vspace{-2mm}
  \includegraphics[width=1\linewidth]{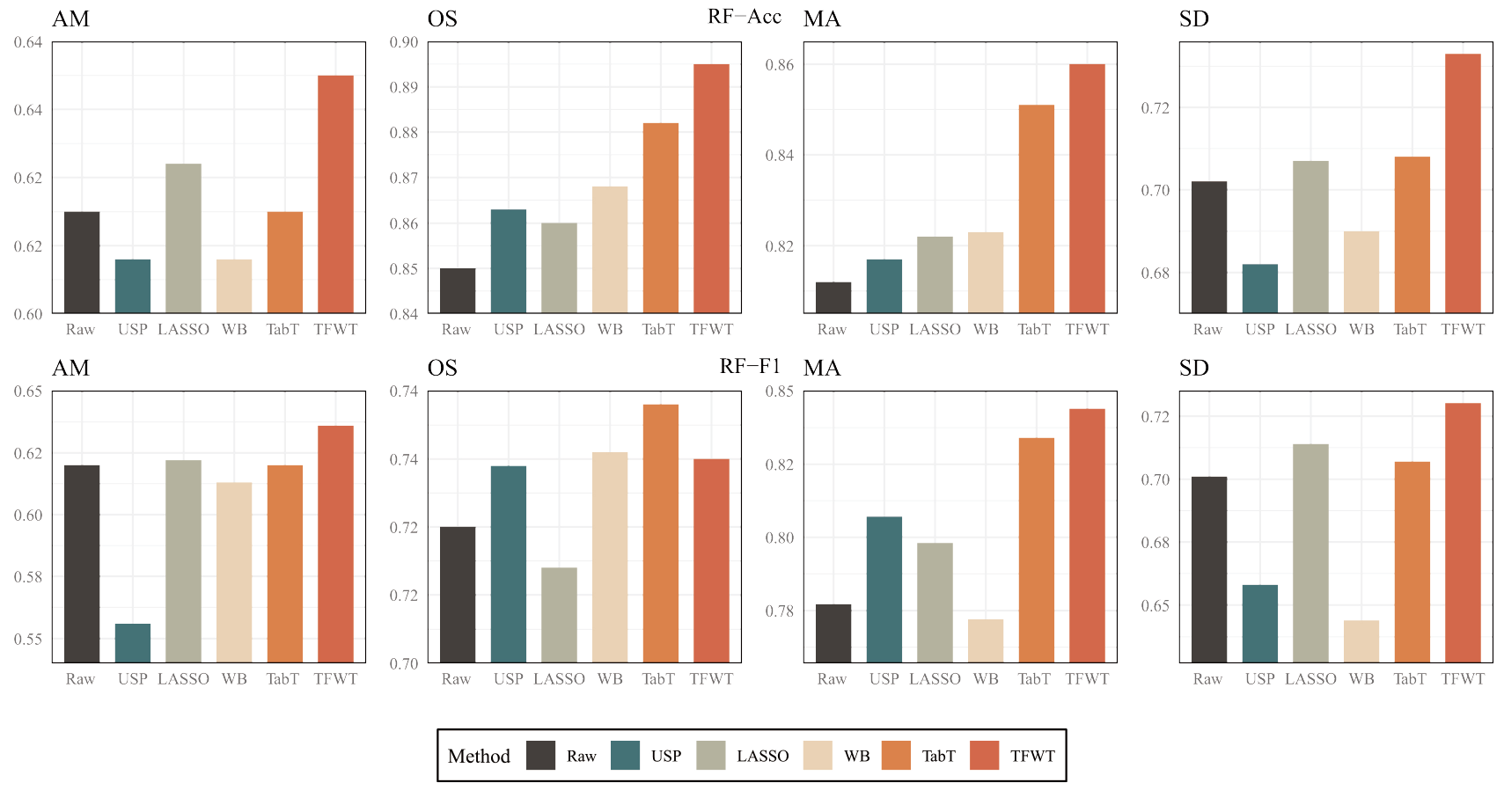}
  \caption{Comparison on RF (Accuracy and F1).}
  \label{fig:mlp-f1}
  % \vspace{-3mm}
\end{figure}

% \begin{figure}[h]
%   \centering
%   \includegraphics[width=0.8\linewidth]{figure/NB-f1.eps}
%   \caption{Comparison on NB (F1 Score)}
%   \label{fig:nb-f1}
% \end{figure}

\noindent \textbf{Superiority over Baseline Models.} We examine the impacts of our TFWT approach and conduct comparative analyses against four baseline models across four performance metrics. 
Our primary metric of representation is overall accuracy, depicted in Figure~\ref{fig:Accuracy Improvement Comparison}. Our TFWT method consistently achieves the highest accuracy on all four datasets. Particularly noteworthy is the comparison with TabTransformer model. While TabTransformer also integrates a Transformer structure in the feature preprocess, TFWT demonstrates a marked superiority in accuracy and F1. 

% \noindent \textbf{Superiority Over Baseline Models.} We assess the effectiveness of the TFWT method by conducting comprehensive comparative evaltuions against four established baseline models. Our primary focus is on overall accuracy, as depicted in Figure~\ref{fig:Accuracy Improvement Comparison}. 

\begin{figure}[thbp!]
  % \vspace{-3mm}

  \centering
  \begin{minipage}[b]{0.45\linewidth}
    \centering
    \scalebox{0.8}{ % Adjust the scale factor to shrink the table
    \begin{tabular}{l|c}
      \hline
      \rule{0pt}{10pt}
      \textbf{Model Name} & \textbf{Mean AUC} \\
      \hline
      USP                 & 0.678 $\pm$ 0.025 \\
      LASSO                & 0.697 $\pm$ 0.020 \\
      WB                & 0.686 $\pm$ 0.014 \\
      TabT                 & 0.697 $\pm$ 0.020 \\
      TFWT           & 0.713 $\pm$ 0.019 \\
      
      \hline
    \end{tabular}
    }
    \captionof{table}{Comparison of Mean AUC.}
    \label{table:models}
  \end{minipage}%
  \hspace{5mm} % You can adjust this space
  \begin{minipage}[b]{0.48\linewidth}
    \centering
    \includegraphics[width=\linewidth]{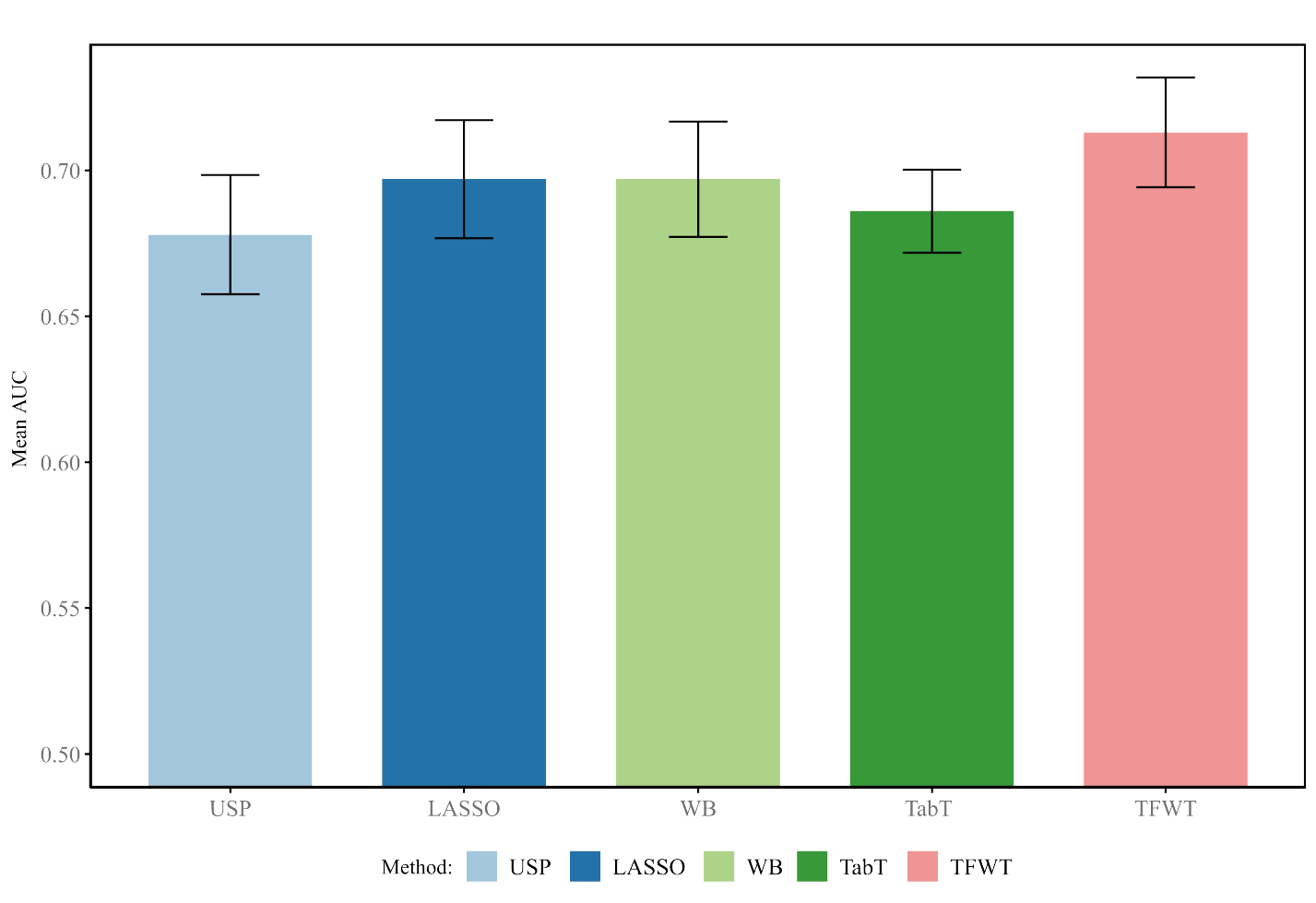}
    \captionof{figure}{Mean AUC of Models.}
    \label{fig:mean-auc}
  \end{minipage}
  % \vspace{-3mm}
  
\end{figure}

% \noindent \textbf{Study of PPO Optimization.} To further enhance performance, we integrate the TFWT pretrained model with PPO optimization techniques. This approach focuses on fine-tuning the model to adapt more effectively to the nuances of various datasets, thus improving its overall predictive accuracy and robustness in downstream tasks.

% \par To quantitatively demonstrate these enhancements, we conduct rigorous experiments across multiple datasets. The results, illustrated in Figure~\ref{fig:exp-3-acc} and Figure~\ref{fig:exp-3-2}, clearly indicate a consistent improvement in performance metrics like accuracy, precision, recall, and F1 score with optimization, compared to the baseline TFWT model without optimization.

\noindent \textbf{Superiority of Fine-Tuning Method.} We further evaluate the advantages brought by our fine-tuning methodology. The refined results post fine-tuning not only match but in several instances surpass the outcomes obtained without fine-tuning, across all evaluated metrics. The key aspect is that the fine-tuned model demonstrates its superiority in the significant reduction of variance across these metrics. Taking Random Forests as a specific example, in a series of five repeated experiments, the variance in results after fine-tuning decreased by $5\%$ to $11\%$ across all four metrics compared with the non-fine-tuned TFWT model. Furthermore, the variance decreased by $7\%$ to $13\%$ compared with raw random forest.

% \begin{figure}[h]
%   \centering
%   \includegraphics[width=1\linewidth]{figure/Exp-3-Acc-3-1.eps}
%   \caption{Comparison on Accuracy}
%   \label{fig:exp-3-acc}
% \end{figure}
\begin{figure}[h]
  \centering
  % \vspace{-1mm}
  \includegraphics[width=0.92\linewidth]{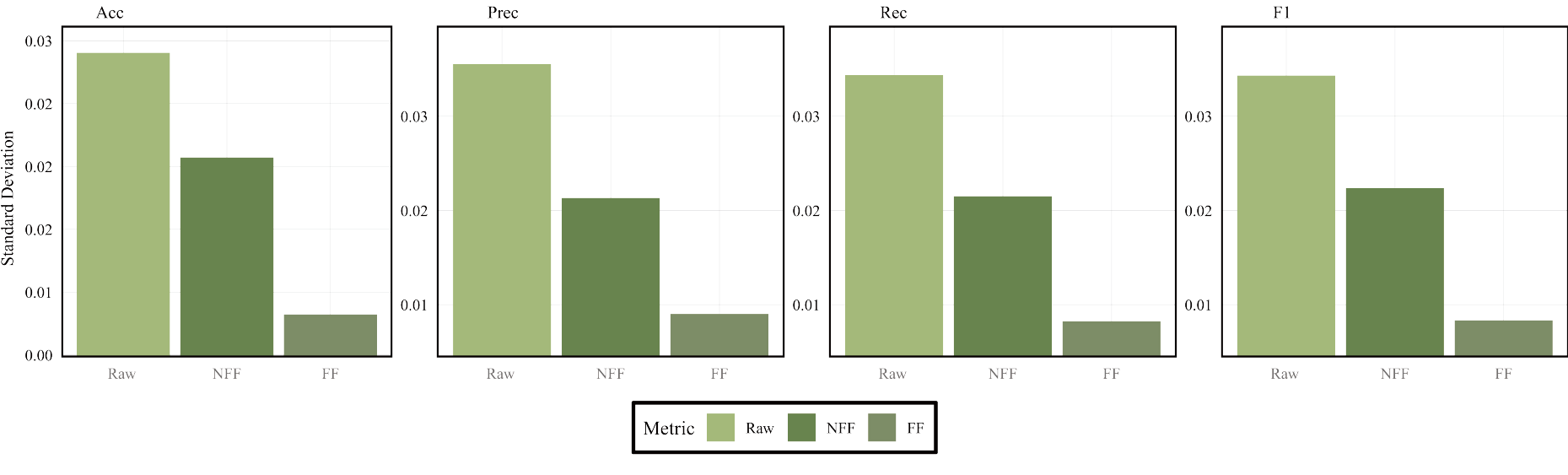}
  \caption{Standard Deviation of Metrics on Random Forest. Here NFF stands for non-fine-tuned and FF stands for fine-tuned.}
  \label{fig:exp-3-acc}
  
\end{figure}
% \begin{figure}[h]
%   \centering
%   \includegraphics[width=0.95\linewidth]{figure/PPO3D-5.jpg}
%   \caption{Comparison on Metrics.}
%   \label{fig:exp-3-2}
%   % \vspace{-3mm}
% \end{figure}

% \input{chapters/5_experiment}
% \input{chapters/6_conclusion}

% \vspace{-7mm}
\section{Conclusion}
% In this study, we present a tabular feature weighting with transformer fr
In this study, we introduce TFWT, a weighting framework designed to automatically assign weights to features in tabular datasets to improve classification performance. Through this method, we utilize the attention mechanism of Transformers to capture dependencies between features to assign and adjust weights iteratively according to the feedback of downstream tasks. Moreover, we propose a fine-tuning strategy adopting reinforcement learning to refine the feature weights and to reduce information redundancy. Finally, we present extensive testing across various real-world datasets to validate the effectiveness of TFWT in a broad range of downstream tasks. The experimental results demonstrate that our method significantly outperforms the raw classifiers and baseline models. 
\clearpage

\section{Acknowledgements}
This work was supported by the Strategic Priority Research Program of the Chinese Academy of Sciences XDB38030300, the National Key R\&D Program of China under Grand No. 2021YFE0108400, the Natural Science Foundation of China under Grant No. 61836013, and Informatization Plan of Chinese Academy of Sciences (CAS-WX2023ZX01-11).
\bibliographystyle{named}
\bibliography{ijcai24}

\end{document}